\documentclass[preprint,12pt]{elsarticle}
\usepackage{latexsym, graphicx, epsfig, amsmath, amssymb, amsfonts}
\usepackage{amsmath}
\usepackage{amssymb}
\usepackage{graphicx}
\usepackage{subfig}
\usepackage{tabu}
\usepackage{adjustbox}
\usepackage{bbold}
\usepackage{bm}
\usepackage{textcomp}
\usepackage[thinc]{esdiff}
\usepackage[dvipsnames]{xcolor}
\usepackage{comment}

\journal{Journal Name}
\begin{document}
\begin{frontmatter}
\title{Physics-Informed Neural Networks for Nonhomogeneous Material Identification in Elasticity Imaging}

\author{Enrui Zhang\textsuperscript{a}}
\author{Minglang Yin\textsuperscript{bc}}
\author{George Em Karniadakis\textsuperscript{a}\corref{cor1}}
\cortext[cor1]{Corresponding author: george\_karniadakis@brown.edu}
\address[a]{Division of Applied Mathematics, Brown University, Providence, RI 02912}
\address[b]{Center for Biomedical Engineering, Brown University, Providence, RI 02912}
\address[c]{School of Engineering, Brown University, Providence, RI 02912}


\begin{abstract}
We apply Physics-Informed Neural Networks (PINNs) for solving identification problems of nonhomogeneous materials. We focus on the problem with a background in elasticity imaging, where one seeks to identify the nonhomogeneous mechanical properties of soft tissue based on the full-field displacement measurements under quasi-static loading. In our model, we apply two independent neural networks, one for approximating the solution of the corresponding forward problem, and the other for approximating the unknown material parameter field. As a proof of concept, we validate our model on a prototypical plane strain problem for incompressible hyperelastic tissue. The results show that the PINNs are effective in accurately recovering the unknown distribution of mechanical properties. By employing two neural networks in our model, we extend the capability of material identification of PINNs to include nonhomogeneous material parameter fields, which enables more flexibility of PINNs in representing complex material properties.
\end{abstract}

\end{frontmatter}


\section{Introduction}
In recent years, we have witnessed the rapid development of deep learning algorithms and their promising applications in computer science \cite{goodfellow2016deep,lecun2015deep} and various fields in physical sciences, biological sciences and engineering \cite{oishi2017computational,bock2019review,brunton2020machine,mamoshina2016applications,elton2019deep}. Typically, deep learning models are data-hungry as they rely on a large amount of training data as the guidance. As deep learning models are applied in physical sciences, the knowledge of physics is implicitly embedded into the models through training data, which is governed by the underlying physical laws. Effective approaches, however, seek more explicit and efficient ways to encode physical knowledge into deep learning models. Recently, the Physics-Informed Neural Networks (PINNs) \cite{raissi2019physics} were proposed as a deep learning framework for solving forward and inverse problems involving partial differential equations (PDEs). For physical systems described mathematically by PDEs, PINNs can explicitly incorporate the underlying physics through the embedding of PDEs. PINNs have been applied for solving forward and inverse problems in fluid mechanics \cite{raissi2020hidden}, solid mechanics \cite{haghighat2020deep}, biomaterials \cite{yin2020non}, optics \cite{chen2020physics} and so on. Other variants of PINNs were proposed for solving fractional PDEs \cite{pang2019fpinns}, stochastic PDEs \cite{yang2018physics}, and for introducing uncertainty by combining Bayesian neural networks \cite{yang2020b}. Through these works, PINNs have been proven to be successful in integrating physics in the framework of deep learning, achieving the concurrent utilization of physics as explicit knowledge and data as implicit knowledge.

In this paper, we extend PINNs to solve inverse identification problems of nonhomogeneous materials in continuum solid mechanics. Specifically, we consider the biomedical scenario of the elasticity imaging problem \cite{doyley2012model}: given the measured full-field displacement data of soft tissue under quasi-static loading, we seek to identify the distribution of mechanical properties of the tissue. Such a procedure can help distinguish normal and diseased tissues. In Method, we formulate the PINN with governing PDEs of hyperelastic solids encoded as the physics. To tackle the spatial dependence of mechanical properties, we introduce two independent neural networks in the PINN: one for approximating the distribution of material parameters, and the other for approximating the solution fields of the corresponding forward problem as in the standard formulation of PINNs. In Computational Tests, we validate the performance of the PINN in identifying the unknown shear modulus field with a prototypical plane strain problem for incompressible, hyperelastic tissue as a proof of concept. Finally, a short summary is provided in Concluding Remarks.

\section{Method}

\begin{figure}[t]
    \centering
    \includegraphics[width=0.7\textwidth]{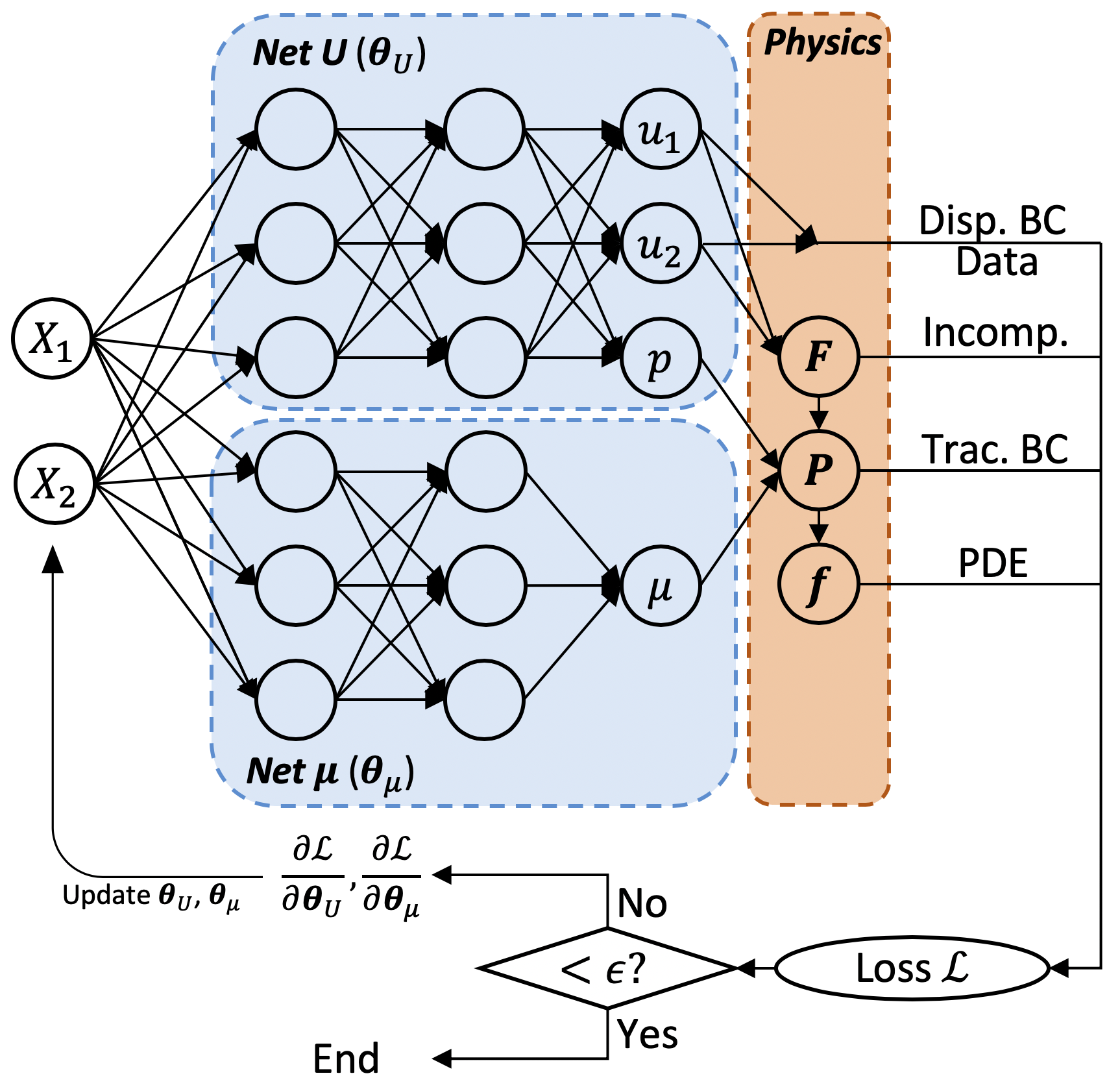}
    \caption{Architecture of the PINN for inverse identification of nonhomogeneous materials. The setup is based on plane strain problems for incompressible Neo-Hookean materials. The loss function $\mathcal{L}$ is formulated according to displacement boundary conditions, data, the incompressibility constraint, traction boundary conditions, and PDEs.}
    \label{fig:Schematics}
\end{figure}

In this part, we formulate the PINN for inverse problems in nonhomogeneous hyperelastic solids (see Fig. \ref{fig:Schematics}). Our aim is to infer the distribution of material parameters according to observed displacement data. The architecture here is based on plane strain problems for incompressible Neo-Hookean hyperelastic materials, where the only one independent material parameter is the shear modulus $\mu$.

We define the plane strain problem in the $\mathbf{X}=(X_1,X_2)$ plane with the Lagrangian description. Following the standard formulation of PINNs, we apply a neural network (\textit{Net $U$}) to approximate the solution of the corresponding forward problem: 
\begin{equation}
(\hat{\mathbf{u}},\hat{p})=(\hat{u}_1,\hat{u}_2,\hat{p})=\mathcal{NN}_U(X_1,X_2;\theta_U),
\end{equation}
where $\theta_U$ stands for the trainable parameters of \textit{Net $U$}. The hat superscript refers to the approximate value by neural networks. $\hat{\mathbf{u}}=(\hat{u}_1,\hat{u}_2)$ is the displacement field with the two components in $X_1$ and $X_2$ directions, respectively. Due to the incompressibility constraint, the pressure field $\hat{p}$ as a Lagrange multiplier is necessary as part of the solution of the forward problem, with which the stress field is uniquely determined. To approximate the spatially dependent shear modulus $\mu$, we apply a second neural network (\textit{Net $\mu$}):  \begin{equation}
\hat{\mu}=\mathcal{NN}_\mu(X_1,X_2;\theta_\mu),
\end{equation}
where the $\theta_\mu$ includes the trainable parameters of \textit{Net $\mu$}.

With the output quantities from the two neural networks, we can further calculate other mechanical quantities in the PINN according to continuum solid mechanics. We write all the equations in their component forms, where the free indices belong to $\{1,2\}$ because of the plane strain setup. For summation on the dummy indices, we follow the Einstein summation convention and omit the summation symbol. The deformation gradient tensor $\mathbf{F}$ is calculated by $\hat{F}_{iJ} = {\partial\hat{u}_i}/{\partial X_J}$, where the partial derivative is computed through automatic differentiation in the PINN. The first Piola-Kirchhoff (PK) stress tensor $\mathbf{P}$ can thereby be calculated by $\hat{P}_{iJ}=-\hat{p}\hat{F}_{iJ}^{-\text{T}}+\hat{\mu}\hat{F}_{iJ}$ according to the stress-strain relationship of hyperelastic materials.

Next, we formulate the PDEs and boundary conditions (BCs) for the PINN. The residuals of the two equilibrium PDEs $\mathbf{f}=(f_1,f_2)$ for solids without body force are $\hat{f}_i={\partial\hat{P}_{iJ}}/{\partial X_J}$. Due to the incompressibility of the material, we also have the constraint $\text{det}(\hat{\mathbf{F}})=1$. We express displacement (Dirichlet) boundary conditions as $\hat{u}_i=\overline{u}_i$ and traction (Neumann) boundary conditions as $\hat{T}_i:=\hat{P}_{iJ}N_J^0=\overline{T}_i$, where the overline refers to prescribed boundary values and $\textbf{N}^0=(N_1^0,N_2^0)$ is the unit outer normal vector on boundaries in the reference configuration.

By combining the displacement data and all the physics information from PDEs, BCs and the incompressibility constraint, the loss function can now be formulated. Suppose we have the measured displacement data $\mathbf{u}^{(i)*}$ on $N_u$ points located at $\mathbf{X}_u^{(i)}$ ($i\in\{1,...,N_u\}$). We place $N_f$, $N_D$, $N_T$ collocation points at $\mathbf{X}_f^{(i)}$ ($i\in\{1,...,N_f\}$), $\mathbf{X}_D^{(i)}$ ($i\in\{1,...,N_D\}$), $\mathbf{X}_T^{(i)}$ ($i\in\{1,...,N_T\}$) for enforcing the equilibrium PDEs and the incompressibility constraint, displacement boundary conditions, and traction boundary conditions in corresponding domains, respectively. Then the loss function can be written as:
\begin{align}
\mathcal{L}(\theta_U,\theta_\mu)&=\frac{w_u}{N_u}\sum_{i=1}^{N_u}\Big|\hat{\mathbf{u}}(\mathbf{X}_u^{(i)};\theta_U)-\mathbf{u}^{(i)*}\Big|^2\nonumber
+\frac{w_f}{N_f}\sum_{i=1}^{N_f}\Big|\hat{\mathbf{f}}(\mathbf{X}_f^{(i)};\theta_U,\theta_\mu)\Big|^2\nonumber\\
&+\frac{w_f}{N_f}\sum_{i=1}^{N_f}\Big[\text{det}\big(\hat{\mathbf{F}}(\mathbf{X}_f^{(i)};\theta_U)\big)-1\Big]^2\nonumber
+\frac{w_D}{N_D}\sum_{i=1}^{N_D}\Big|\hat{\mathbf{u}}(\mathbf{X}_D^{(i)};\theta_U)-\overline{\mathbf{u}}(\mathbf{X}_D^{(i)})\Big|^2\nonumber\\
&+\frac{w_T}{N_T}\sum_{i=1}^{N_T}\Big|\hat{\mathbf{T}}(\mathbf{X}_T^{(i)};\theta_U,\theta_\mu)-\overline{\mathbf{T}}(\mathbf{X}_T^{(i)})\Big|^2,
\end{align}
where the five terms correspond to the loss components of displacement data, PDEs, the incompressibility condition, displacement boundary conditions, and traction boundary conditions, respectively. The coefficients $w$'s with various subscripts refer to the weights of corresponding loss terms.

The construction of the PINN is now complete. The PINN is trained by minimizing the loss function \begin{equation}
(\tilde{\theta}_U,\tilde{\theta}_\mu)=\operatorname*{argmin}_{\theta_U,\theta_\mu}\mathcal{L}(\theta_U,\theta_\mu).
\end{equation}
Upon convergence, the PINN can predict the shear modulus field at an arbitrary point in the solid $\mathbf{X}=(X_1,X_2)$ according to
\begin{equation}
\tilde{\mu}(X_1,X_2)=\mathcal{NN}_\mu(X_1,X_2;\tilde{\theta}_\mu).
\end{equation}

\section{Computational Tests}

\subsection{Problem Setup}

\begin{figure}
    \centering
    \includegraphics[width=0.7\textwidth]{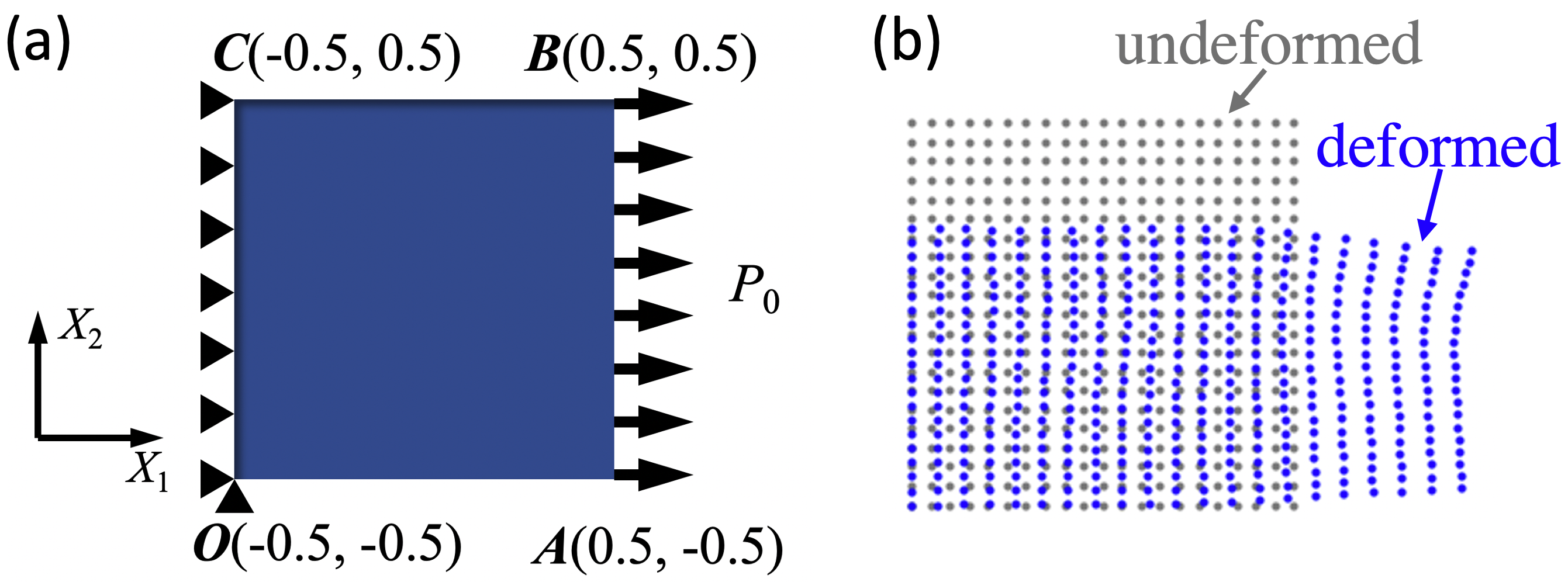}
    \caption{(a) Setup of the prototypical plane strain problem. (b) Spatial distribution of the measurement points in both undeformed and deformed configurations. Displacement data can be calculated by comparing the two configurations.}
    \label{fig:ProblemSetup}
\end{figure}

Here we test the performance of the PINN with a prototypical problem in the background of elasticity imaging. We consider a plane strain problem, which is a common assumption for researches in elasticity imaging \cite{doyley2012model}. See Fig. \ref{fig:ProblemSetup}(a) for the sketch. Consider a square-shaped soft tissue $\bm{OABC}$ in the $(X_1,X_2)$ plane with side length $1.0$. The edges $\bm{OA}$ and $\bm{OC}$ are aligned along $X_1$ and $X_2$ directions, respectively. The tissue is modeled as a nonhomogeneous, incompressible Neo-Hookean material with shear modulus distribution 
\begin{align}
\label{eqn:TrueMu}
\mu^*(X_1,X_2)&=0.333-0.05\big[(X_1+1)^2+(X_2+0.5)^2\big]\nonumber\\
&+0.133\exp{\big\{-22.22[(X_1-0.1)^2+(X_2-0.2)^2]\big\}},
\end{align}
where the second term mimics the normal tissue with a low-frequency, low-magnitude fluctuation of shear modulus and third term mimics the local diseased tissue with abnormally high shear modulus around $(0.1,0.2)$. The left edge $\bm{OC}$ is fixed in $X_1$ direction and the left bottom corner point $\bm{O}$ is fixed in $X_2$ direction. Uniform tensile loading with magnitude $P_0=0.3$ (in the first Piola-Kirchhoff definition) is applied on the right edge $\bm{AB}$. To obtain the displacement data, $21\times21$ measurement points are placed on uniform grid nodes in the square domain in the undeformed configuration (see the points for the undeformed configuration in Fig. \ref{fig:ProblemSetup}(b)). Under the loading, the locations of the points in the deformed configuration are measured (see the points for the deformed configuration in Fig. \ref{fig:ProblemSetup}(b)), hence obtaining the displacement data. To generate the displacement data for the PINN, we employ the Abaqus software to solve the forward boundary value problem in Fig. \ref{fig:ProblemSetup}(a) with the information of true shear modulus distribution in Eqn.~\ref{eqn:TrueMu}.

\subsection{Technical Details}
We formulate the PINN according to the setup of the prototypical problem. Both \textit{Net $U$} and \textit{Net $\mu$} have 4 hidden layers with 30 neurons for each layer. The learning rate is set to be 0.001. Other technical details include the usage of Xavier initialization \cite{glorot2010understanding}, Adam optimizer \cite{kingma2014adam}, and layer-wise ``tanh" adaptive activation function \cite{jagtap2020adaptive}. We place $41\times41$ training points on uniform grid nodes in the square domain in the undeformed configuration, which are used for evaluating the loss terms for PDEs and the incompressibility condition. Each edge is uniformly equipped with $40$ training points for evaluating the loss for displacement (left edge) or traction (right, top, and bottom edges) BCs. The loss for displacement data is evaluated at the $21\times 21$ measurement points. To formulate the loss function in Eqn.~\ref{eqn:TrueMu} with the five loss terms, the weights are set as $w_u=10.0$, $w_T=3.0$, and $w_f=w_D=1.0$. The PINN is trained over 2M ($\text{M}=10^6$) epochs in total.

\subsection{Results}

\begin{figure}
    \centering
    \includegraphics[width=0.7\textwidth]{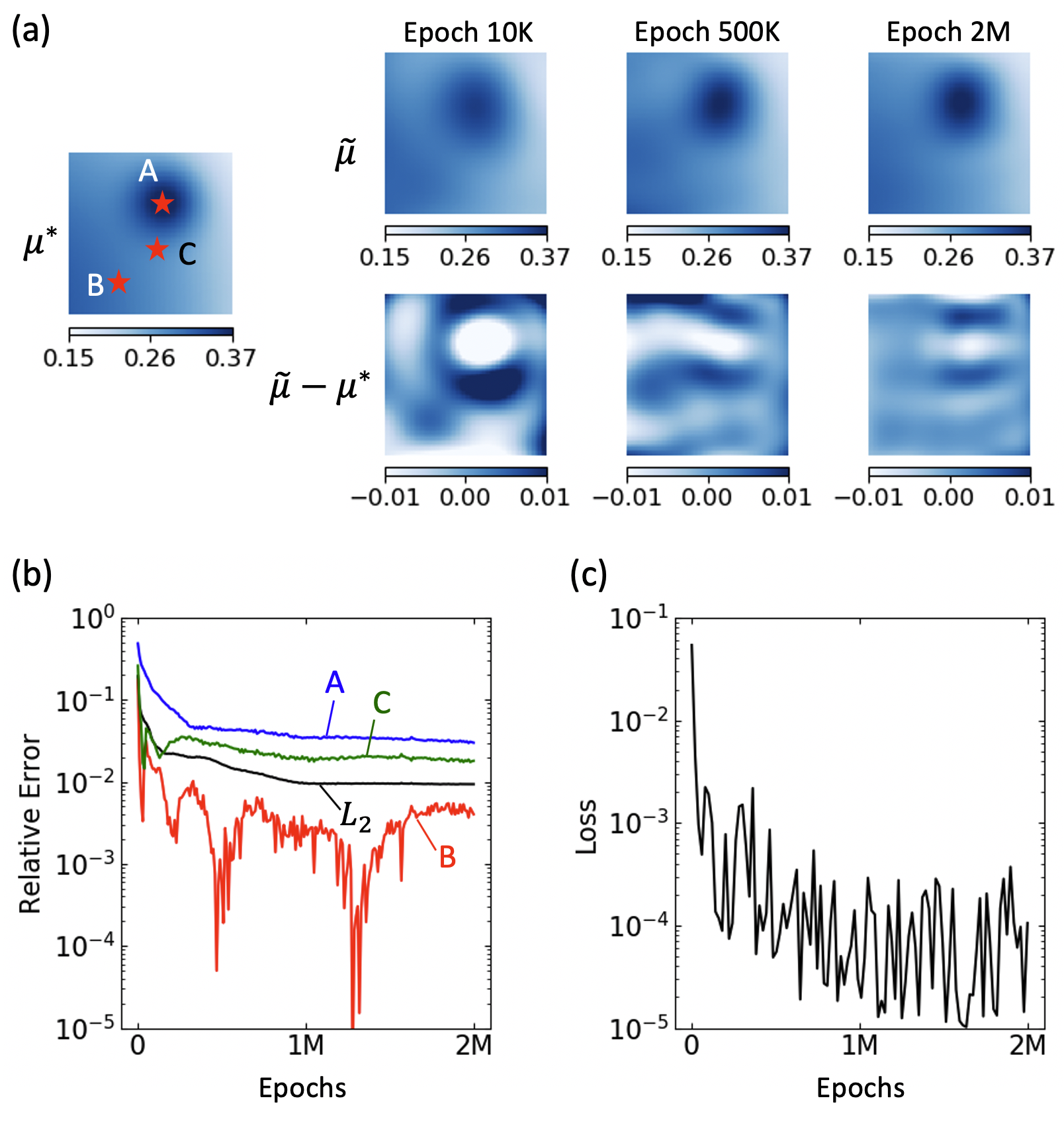}
    \caption{Results of the inverse identification of the nonhomogeneous material. (a) True value $\mu^*$, inferred value $\tilde\mu$, and the signed error $\tilde\mu-\mu^*$ of the shear modulus distribution in the prototypical problem. (b) Evolution of the relative error of the overall modulus field (in the $L^2$ sense) and on points A, B, and C marked in (a). (c) Evolution of the total loss function.}
    \label{fig:PSN_field}
\end{figure}

The results are shown in Fig. \ref{fig:PSN_field}. During the training process, we concurrently test the performance of the PINN by evaluating the predicted shear modulus field. Fig. \ref{fig:PSN_field}(a) displays the true distribution $\mu^*$, the inferred distribution $\tilde{\mu}$, and the inference error $\tilde{\mu}-\mu^*$ after training over 10K ($\text{K}=10^3$), 500K, and 2M epochs. As the training proceeds, the inferred shear modulus field gradually becomes similar to the true pattern. After 2M epochs, the local absolute error is uniformly below $0.01$, compared to the true value $\mu^*$ ranging from roughly $0.15$ to $0.37$. Fig. \ref{fig:PSN_field}(b) shows the evolution of the relative $L^2$ error of the overall modulus field and the relative error at points A, B, and C (marked in Fig. \ref{fig:PSN_field}(a)), normalized by the mean of true modulus field $0.28$. The overall error after 2M epochs is as small as $1\%$. Fig. \ref{fig:PSN_field}(c) shows the evolution of the total loss against the training epochs. The loss gradually decreases from $\mathcal{O}(10^{-1})$ to $\mathcal{O}(10^{-4})$ despite some fluctuations, indicating that the PDEs, BCs, the incompressibility condition, and the displacement data are all approximately satisfied. Combining all the results shown in Fig. \ref{fig:PSN_field}, the PINN has sufficiently utilized the conditions from physics and data and accurately recovered the shear modulus field in our prototypical problem.

\section{Concluding Remarks}
We have demonstrated the capability of PINNs in accurately solving inverse identification problems of nonhomogeneous hyperelastic materials. Inheriting the merits of the standard formulation of PINNs, our model only requires a small amount of data for the current problem, compared to typical deep learning models requiring big data for training. By introducing the neural network for the approximation of the material parameter field, we extend the capability of PINNs for inverse problems from inferring constant material parameters to spatially dependent material parameters, achieving greater flexibility in describing material properties. Such a framework broadens the application scenarios of PINNs to more practical problems including, in our case, the elasticity imaging for soft tissues. It is worth noting that our prototypical problem serves as a simple proof of concept for the practicability of PINNs for spatially dependent material properties. Further study should include a variety of factors that may influence the validity and accuracy of the inference results of PINNs, such as the maximum-to-minimum ratio and the spatial frequency of the true modulus, performance for hyperelastic constitutive models with multiple parameter fields, model robustness to noisy data and multiple loading conditions.

\section*{Acknowledgment}
The work is supported by grant U01 HL142518 of National Institute of Health.

\bibliographystyle{elsarticle-num-names}
\bibliography{reference}

\end{document}